\documentclass{article}

\usepackage{PRIMEarxiv}

\usepackage[utf8]{inputenc} 
\usepackage[T1]{fontenc}    
\usepackage{hyperref}       
\usepackage{url}            
\usepackage{booktabs}       
\usepackage{amsfonts}       
\usepackage{nicefrac}       
\usepackage{microtype}      
\usepackage{lipsum}
\usepackage{fancyhdr}       
\usepackage{graphicx}       
\usepackage{multirow}
\graphicspath{{media/}}     

\title{Swinv2-Imagen: Hierarchical Vision Transformer Diffusion Models
for Text-to-Image Generation
}

\author{
  Ruijun Li \\
  Auckland University of Technology \\
  Auckland, New Zealand\\
  \texttt{zjc0233@autuni.ac.nz} \\
  \And
  Weihua Li \\
  Auckland University of Technology \\
  Auckland, New Zealand\\
  \texttt{weihua.li@aut.ac.nz} \\
  \And
  Yi Yang \\
  Hefei University of Technology \\
  Hefei, China \\
  \texttt{yyang@hfut.edu.cn} \\
  \And
  Hanyu Wei \\
  University of Tasmania \\
  Hobart, Australia \\
  \texttt{hanyu.wei@utas.edu.au} \\
  \And
  Jianhua Jiang \\
  Jilin University of Finance and Economics \\
  Changchun, China \\
  \texttt{jjh@jlufe.edu.cn} \\
  \And
  Quan Bai \\
  University of Tasmania \\
  Hobart, Australia \\
  \texttt{quan.bai@utas.edu.au} \\
}

\begin{document}
\maketitle

\begin{abstract}
Recently, diffusion models have been proven to perform remarkably well in text-to-image synthesis tasks in a number of studies, immediately presenting new study opportunities for image generation. Google's Imagen follows this research trend and outperforms DALLE2 as the best model for text-to-image generation. However, Imagen merely uses a T5 language model for text processing, which cannot ensure 
learning the semantic information of the text. Furthermore, the Efficient UNet leveraged by Imagen is not the best choice in image processing. To address these issues, we propose the Swinv2-Imagen, a novel text-to-image diffusion model based on a Hierarchical Visual Transformer and a Scene Graph incorporating a semantic layout. In the proposed model, the feature vectors of entities and relationships are extracted and involved in the diffusion model, effectively improving the quality of generated images. On top of that, we also introduce a Swin-Transformer-based UNet architecture, called Swinv2-Unet, which can address the problems stemming from the CNN convolution operations. Extensive experiments are conducted to evaluate the performance of the proposed model by using three real-world datasets, i.e., MSCOCO, CUB and MM-CelebA-HQ. The experimental results show that the proposed Swinv2-Imagen model outperforms several popular state-of-the-art methods.
\end{abstract}


\section{Introduction}
People tend to describe rich and detailed pictures of scenes through language, and the ability to generate images from these descriptions can facilitate creative applications in various life contexts, including art design and multimedia content creation \cite{kim2020tivgan,li2020exploring}. This fact has inspired researchers to design  models of text to image comparative learning to assist people with making decisions quickly in specific scenarios, such as presentation and advertising design \cite{mathesul2021attngan,Park2021BenchmarkFC}. In recent years, diffusion models have attracted the attention of many scholars due to their promising performance in image generation. Within this framework, DALL-E 2 \cite{Ramesh2022HierarchicalTI} and Imagen \cite{Saharia2022PhotorealisticTD} have become successful generative models for image generation. 

Imagen is currently one of the greatest image generation models. Its most significant distinguishing feature is its immensity, which is reflected, in particular, by its utilisation of a large text encoder,  i.e., T5 \cite{Raffel2020ExploringTL}. T5 is pre-trained on a sizable plain text corpus. It turns out that T5 is very effective for enhancing image fidelity and image-text alignment \cite{Saharia2022PhotorealisticTD}. However, T5 is Transformer-based, and using T5 alone to obtain text embeddings cannot guarantee that the model learns important text features, such as semantic layout. Besides visual elements, the semantic layout is recognised as an important factor in guiding text-to-image synthesis \cite{Li2019ObjectDrivenTS}. Our experimental results provide evidence for this claim. 

Furthermore, very few research works are dedicated to addressing the UNet issue of Imagen. The diffusion model of Imagen relies on the Efficient-UNet, which suffers from the limitations of CNN convolution operations. CNN are good at extracting the low-level features and elements of visual structure, such as colour, contour, texture and shape \cite{ganar2014enhancement}. However, CNN just focuses on the consistency of these low-level features under transformations, such as translation and rotation, i.e, Translation Invariance \cite{kauderer2017quantifying} and Rotation Invariance \cite{chidester2018rotation}. It fails to extract features efficiently in terms of global and layout. In other words, the convolutional filter detects key points, object boundaries and other basic units that constitute the visual elements. The elements should be transformed simultaneously under translation and rotation transformations. Therefore, CNN is one of the choices for addressing the consistency problem, though it cannot represent high-level visual semantic information regarding the connections between objects. For text-to-image synthesis tasks, it is significant to consider how such basic elements can be organised as objects and how they can be constructed into a scene using spatial location relationships. The Transformer is more natural and efficient than CNN in processing this high-level semantic information. This is mainly because the Transformer focuses on local information and has a diffusion mechanism to find expressions from the local to global layout \cite{li2021local,liang2022local}. 

To solve the aforementioned drawbacks of Imagen, in this paper, we propose a diffusion text-to-image generation model called Swinv2-Imagen. The proposed model is based on a Hierarchical Visual Transformer and Scene Graph incorporating layout information. Specifically, the semantic layout is generated via semantic scene graphs, enabling Swinv2-Imagen to parse the layout information in the text description effectively. In this paper, we adopt Stanford Scene Graph Parser \cite{Johnson2018ImageGF} to obtain the Scene Graph from the text. Subsequently, the entity and relationship embeddings are extracted using a frozen Graph Convolution Network (GCN) \cite{Johnson2018ImageGF}. The image generation process appears conditional on text, object and relationship embeddings. The layout representation with global semantic information ensures the realism of the generated images. In addition, the diffusion models are developed based on Swinv2-Unet, a variant of Swin Transformer v2 \cite{Liu2022SwinTV}, which implements a full Transformer text to image generation model for a more accurate understanding of text semantics. Finally, we evaluate our model on the MSCOCO, CUB and MM-CelebA-HQ datasets. The results show that the proposed model outperforms the current best generative model, Imagen, on MSCOCO. The ablation experiments reveal that the addition of semantic layouts is effective in improving the semantic understanding of the model.

The key contributions of this paper are summarised below.
\begin{enumerate}
    
    \item We leverage scene graphs to extract entity and relational embeddings to improve local and layout information representation of text for a more accurate understanding of the text and realistic image generation;
    \item We propose Swinv2-UNet as a novel diffusion model architecture to implement a full Transformer text generation image model; 
    \item We combine the scene graph with the Transformer to improve the effectiveness of the model;
    \item We achieve a new state-of-the-art FID result, (FID=7.21), on the MSCOCO dataset compared to the latest generative models. Better results are also obtained on both the CUB (FID=9.78) and MM CelebA-HQ (FID=10.31).

\end{enumerate}

The rest of the paper is organised as follows. In Section 2, related works are reviewed. In Section 3, we elaborate on the proposed Swinv2-Imagen model. In Section 4, we conduct extensive experiments to evaluate the performance of the proposed model and perform an ablation study to evaluate the contributions of each key component of our model. Finally, we conclude this paper in Section 5, and discuss future research directions.

\section{Related work}
\subsection{Diffusion models}
Text-to-Image synthesis is a typical application of multimodal and cross-modal comparative learning. In the field of image generation, most models mainly fall into two categories, i.e., the GAN-based generation models \cite{Zhu2019DMGANDM, Zhu2020CookGANCB, Zhang2019StackGANRI, Xia2021TediGANTD, Crowson2022VQGANCLIPOD, Cheng2020RiFeGANRF} and the diffusion-based models \cite{Ho2020DenoisingDP, Ho2022CascadedDM,Nichol2022GLIDETP, Rombach2022HighResolutionIS,Song2021DenoisingDI}. The former has been developed over the last few years and widely used in many scenarios, such as medical and image restoration. The latter has demonstrated outstanding performance over the GAN models, acknowledged as state-of-the-art deep generative models \cite{Saharia2022PhotorealisticTD, Dhariwal2021DiffusionMB, Yang2022DiffusionMA}.

Diffusion models and GAN generative models are essentially comparable, both being a process of gradually removing noise. However, in contrast to GAN, the diffusion models do not suffer from training instability and model collapse. The diffusion model transforms the data distribution into random noise and reconstructs data samples with the same distribution \cite{Saharia2022PhotorealisticTD,Cao2022ASO}. The diffusion model demonstrates outstanding performance for a number of tasks, such as multimodal modelling. Many contemporary text-to-image synthesis models, e.g., DALL-E 2 \cite{Ramesh2022HierarchicalTI}, Imagen \cite{Saharia2022PhotorealisticTD} and GLID \cite{Nichol2022GLIDETP}, are constructed based on the diffusion model. They cascade multiple diffusion models to improve the image generation quality step by step. DALL-E 2 uses a priori diffusion model and CLIP Latents to process the text. In contrast, Imagen discards the priori model and replaces it with a large pre-trained text encoder, i.e., T5. Although the T5 model leveraged in the Imagen model improves the understanding of the text, it does not ensure that the model understands the semantic layout of the text, especially in complex sentences containing multiple objects and relationships. As a result, the model will not be able to reproduce some entities or will lose some entity relationships. Therefore, we attempt to model the global semantic layout by adding a scene graph in the text processing. Furthermore, Imagen builds its diffusion model based on Efficient-Unet. Efficient-Unet is not the best choice in image generation tasks, because it contains multiple CNN blocks and leads to a limited view within the CNN kernel window. 

\subsection{Scene graph and Graph representation learning}
A sentence’s nature is a linear data structure, where one word follows another \cite{Johnson2018ImageGF}. Usually, when a sentence is complex with multiple objects, it is time-consuming to analyse the sentence directly, and the accuracy of the text-image alignment is not guaranteed. Complex sentences often incorporate rich scene information. Mapping this information into a scene graph can provide an intuitive understanding of the relationships between objects in a sentence \cite{mittal2019interactive}. Previous studies reveal that the performance of multimodal models, such as text-to-image synthesis, is significantly dependent on mining visual relationships \cite{Zhu2022SceneGG}. Scene graphs can provide a high level of understanding regarding scene information \cite{Johnson2018ImageGF}. Therefore, the scene graph is recognised as a useful representation of images and text. Specifically, each node in a scene graph represents an object, such as a person or an event, and each object has multiple attributes, such as shape. The relationships between objects are denoted by the edges between nodes, which can be an action or a position \cite{Chang2021ACS}. Recently, the scene graphs have been used extensively for tasks such as text-based image retrieval \cite{Johnson2015ImageRU, Schuster2015GeneratingSP}, semantic segmentation \cite{Taghanaki2020DeepSS, Jaritz2020xMUDACU}, visual question answering \cite{Li2019RelationAwareGA}, image captioning \cite{Gao2018ImageCW, Yang2019AutoEncodingSG, Zhong2020ComprehensiveIC, Gu2019UnpairedIC} and image generation \cite{Johnson2018ImageGF, mittal2019interactive, li2019pastegan, zhao2019image}. 

In addition, there is no way for an image generation model to manipulate graph-like data such as scene graphs directly, so scene graphs are usually used in conjunction with graph representation learning \cite{Hamilton2020GraphRL}. The main objective of graph representation learning is to extract node and edge contexts from the scene graph and map them to a set of embeddings.  Graph representation learning methods can currently be classified into two types, i.e., machine learning based on Random-Walk and deep learning Graph Convolution-based methods \cite{Hamilton2020GraphRL}. Node2vec \cite{Grover2016node2vecSF} is a typical representative model of the former. It is based on SkipGram \cite{Mikolov2013EfficientEO} theory to learn the embedding of nodes on a graph and optimises the sampling method. It is proposed in related studies \cite{Chen2020GraphRL} that two sampling methods, Breadth-First Search (BFS) and Depth First Search (DFS), are mainly included when sampling neighbouring nodes in a graph. BFS requires that each sampled node is a direct neighbour of that node. This sampling method results in a graph representation that is more concerned with local information. In contrast, DFS, where each node is sampled to increase the distance to the initial node as much as possible, produces a graph representation that focuses more on global information. Random-Walk-based representation learning \cite{Hamilton2017RepresentationLO} is composed of multiple stages, each with different optimisation goals, which is a typical non-end-to-end model. Graph convolution-based methods, e.g. graph convolution neural networks \cite{Johnson2018ImageGF}, are able to learn both node feature information and structural information via an end-to-end way. It focuses on both local information and global structural features. Graph convolution is extremely applicable to nodes and graphs of topology. It is currently the best choice for graph data learning.

\subsection{UNet}
UNet is an encoder-decoder architecture, which is scalable in structure \cite{Ronneberger2015UNetCN}. The encoding stage of the UNet consists of four downsamples. Symmetrically, its decoding stage is also upsampled four times, restoring the result of the encoder to the resolution of the original image. In contrast to Fully Convolutional Networks (FCN) \cite{Shelhamer2015FullyCN}, UNet upsamples four times and uses a jump connection in the encoder and decoder of the corresponding convolution blocks. The jump connection ensures that the final recovered feature map incorporates more low-level semantic features and features at different scales are well fused, allowing for multi-scale prediction. In addition, the four times up-sampling also allows the segmentation map to recover information such as edges more finely. However, UNet also has some shortcomings. For example, UNet++ \cite{Zhou2018UNetAN, Zhou2020UNetRS} argues that it is inappropriate to directly combine the shallow features from the encoder with the deeper features from the decoder in UNet. Direct fusion would potentially lead to semantic gaps. Furthermore, UNet 3+ \cite{Huang2020UNet3A} maximises the scope of model information fusion and circulation. Each decoder layer in the UNet 3+ fuses small-scale and same-scale feature maps from the encoder with larger-scale feature maps from the decoder, which capture both fine-grained and coarse-grained semantics at full scale.

Many researchers develop a set of UNet variants by improving and optimising the original UNet. For example, ResUNet \cite{Zhang2018RoadEB} and DenseUNet \cite{Cai2020DenseUNetAN} are inspired by Residual and Dense connections, respectively; each sub-module of the U-Net is replaced with a form having a Residual connection and a Dense connection. There are variants, e.g., MultiResU-Net \cite{Ibtehaz2020MultiResUNetR} and R2 U-Net \cite{Alom2018RecurrentRC}. All of these models are constructed using multiple convolutional blocks. With the advent of the Transformer, researchers begin to develop the UNet base on the Transformer, such as Swin-UNet \cite{Cao2021SwinUnetUP}. While Swin-UNet mitigates the limitations of CNN convolutional operations, it is likely to suffer from training instability due to the use of the Swin-Transformer block. Swin-Transformer v2 \cite{Liu2022SwinTV} is an improvement on Swin-Transformer, which is effective in avoiding training instability and is easier to scale. 

Inspired by these research works, we propose a Swinv2-Imagen model that leverages scene graphs as auxiliary modules to help the model understand the text semantics more comprehensively. In addition, Swinv2-Unet is applied to build the diffusion models so that our model is based on the full Transformer implementation. As a result, it effectively addresses the limitations of CNN convolution operations, theoretically enabling the synthesis of images better than baselines.


\section{Swinv2-Imagen}

\begin{figure}[!ht]
    \centering
    \includegraphics[width=0.72 \textwidth]{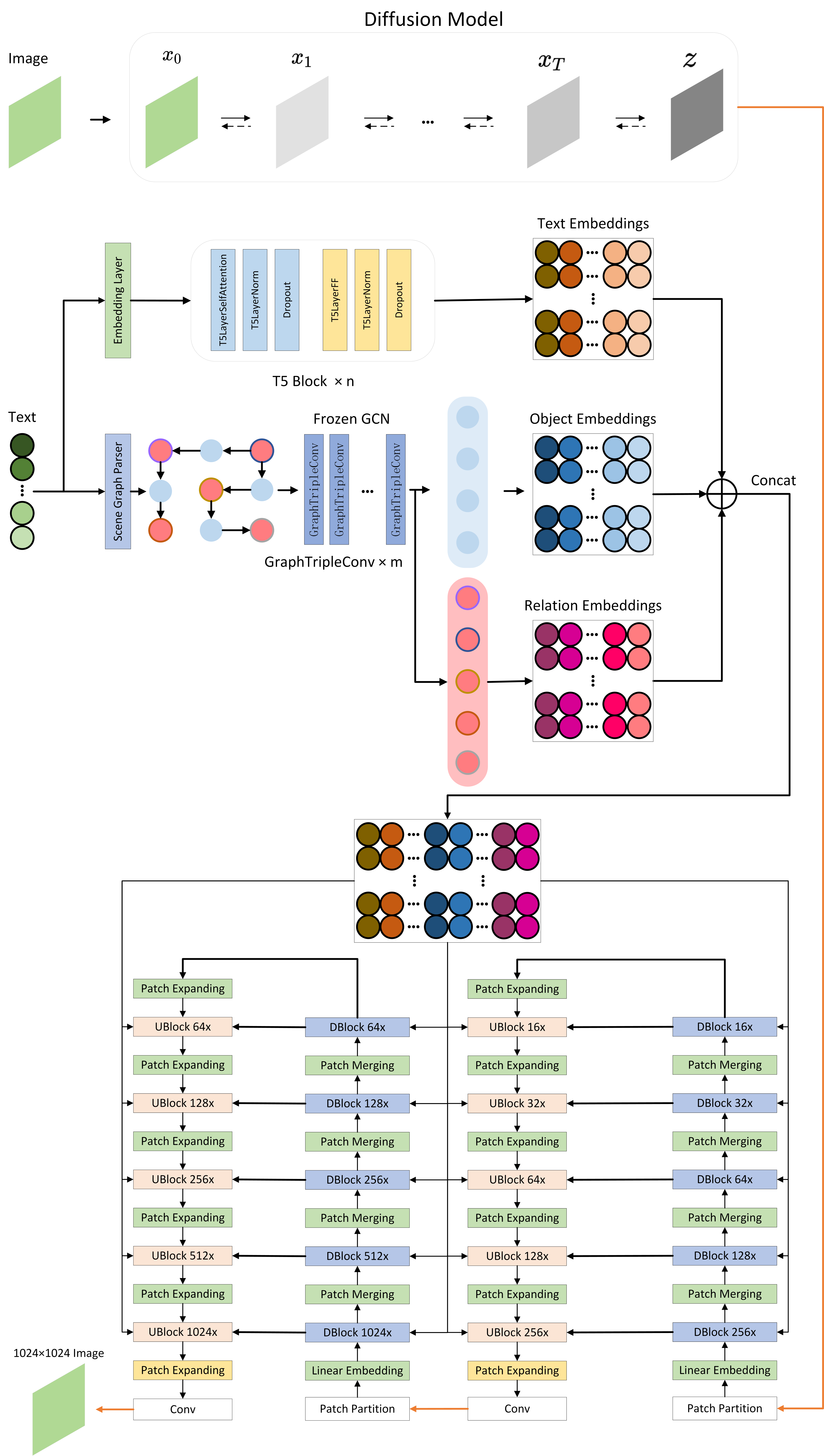}
    \caption{The overall architecture of Swinv2-Imagen.}
    \label{fig:architecture}
\end{figure}

The overall architecture of the proposed Swinv2-Imagen model is shown in Figure \ref{fig:architecture}. It takes text descriptions as input and uses scene graphs to guide downstream image generation more accurately and efficiently. The upstream comprises two sub-modules: the text encoder, which maps the text input to a text embedding sequence and the scene graph generator sub-module. The downstream consists of a set of conditional diffusion models, integrating the intermediate embeddings in the upstream and generating high-fidelity images step by step. The scene graph generator includes a Scene Graph parser and a frozen Graph Neural Network, which aims to represent objects and relationships in a text with a graph structure. 

The input of the model is a text-picture pair. Firstly, the text is encoded by T5 tokenizers and input to the embedding layer to get the initial text embedding. Next, it goes through the T5 encoder (n-layer T5 Block) to obtain Text Embeddings. Meanwhile, the scene graph parser extracts the scene graph from the text, and the frozen GCN (m-layer Graph Triple Convolution) obtains the corresponding Object and Relation embeddings. Finally, the Conditional embeddings are obtained by concatenating the Text embeddings, Object embeddings and Relation embeddings in this order. The Conditional embeddings are used as conditional input for subsequent super-resolution image generation. In the following subsections, we describe the main components of Swinv2-Imagen in detail.

\subsection{Pre-trained frozen text encoders}
It is widely acknowledged that a robust semantic text encoder is essential for text-to-image synthesis models and plays a crucial role in analysing the complexity and composition of textual input \cite{Saharia2022PhotorealisticTD}. Previously, language models were mainly built on RNN architectures. However, since the emergence of the Transformer, a number of transformer-based pre-trained language models have been developed, such as GPT \cite{Radford2018ImprovingLU, Radford2019LanguageMA, Brock2019LargeSG}, BERT \cite{Devlin2019BERTPO} and T5 \cite{Raffel2020ExploringTL}. The traditional Imagen model is compared against popular text encoders, BERT, CLIP and T5-XXX, by freezing parameters. The existing research results prove the promising performance of T5-XXX in terms of both image-text alignment and image fidelity \cite{Saharia2022PhotorealisticTD}. Therefore, we adopt the T5 large language model for text encoding in the proposed model.

\subsection{Scene Graph and Frozen Graph Convolutional Neural Network}

\begin{figure}[!h]
    \centering
    \includegraphics[width=5in]{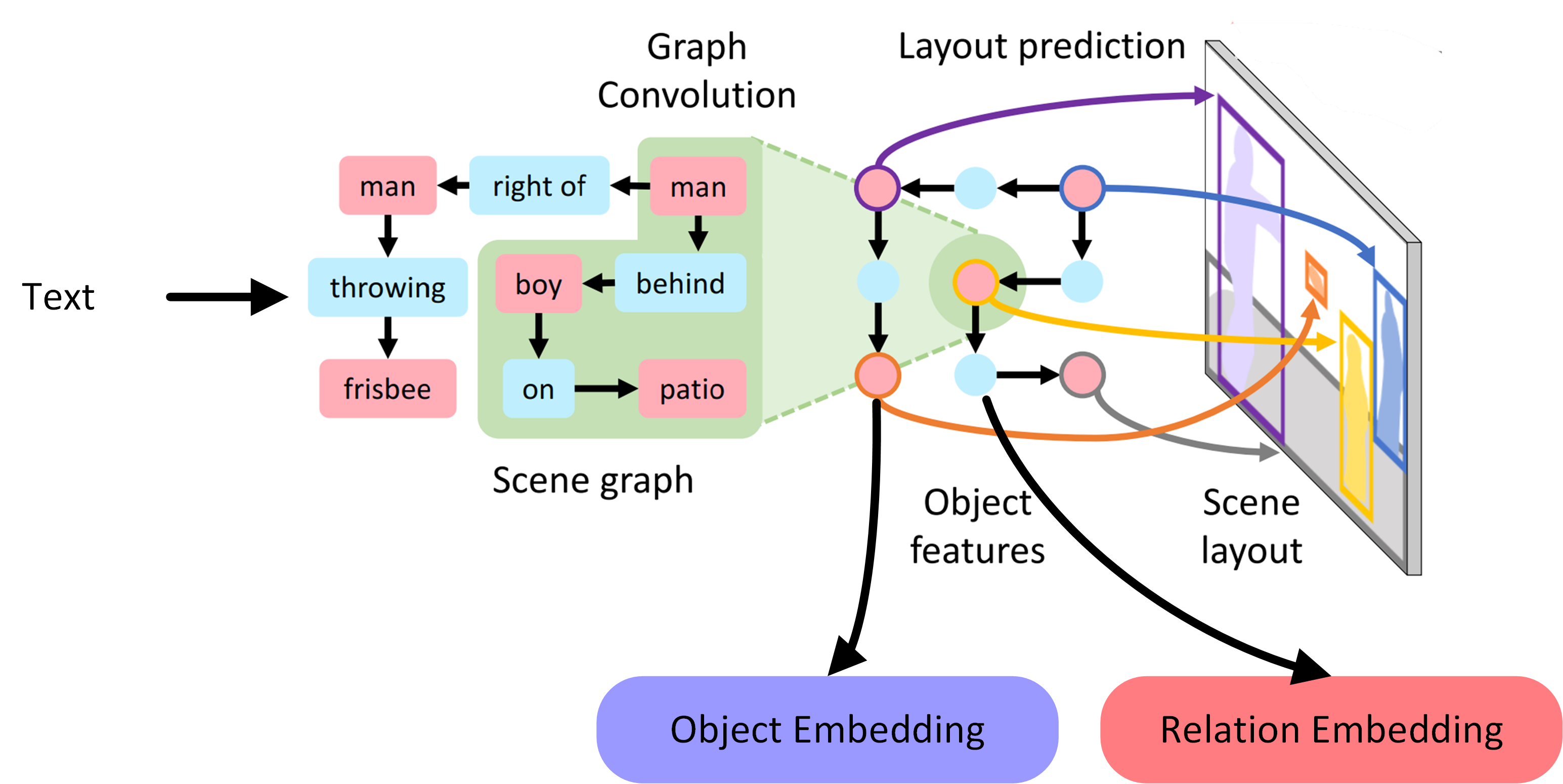}
    \caption{The process of Object and Relation embeddings extraction.}
    \label{fig:sg}
\end{figure}
This sub-module aims to extract entity relationships from the text to enhance the text understanding of the model. We adopt a Scene Graph parser to represent text as a scene graph, followed by a frozen GCN to extract the entity and relational embeddings for the image generation with diffusion models. Scene graphs with graph neural networks \cite{Johnson2018ImageGF} have been proven to be highly effective in extracting object relationships from the text. As shown in Figure \ref{fig:sg}, Swinv2-Imagen constructs a scene graph for the text and is followed by a graph neural network to extract the entities and relationships from the scene graph. For any given text description, the corresponding scene graph is represented as $(O, E)$, where $O=(o_1,o_2,o_3,\cdots,o_n )$   denotes each entity object in the sentence i.e., subject and object, and $E$ is a collection of edges of the form $(o_i, r, o_j)$, where $r \in \mathcal{R} $, $ \mathcal{R} $ refers to a collection of relationships. $E$ denotes the relationship between the two entities, e.g., action and position. In the end, object and relation embeddings are constructed, which are used to assist the T5 model in analysing and understanding the text more comprehensively.

The input to the graph convolution is a scene graph, having each node and edge represented as a vector with dimension $D^{in}$. In the graph convolution model, these vectors are newly computed as new vectors with dimension  $D^{out}$ for each node and edge. In addition, two functions, $g_s$ and $g_o$, are used to calculate the entity features vectors of output. They use a triplet of an edge, $ \left( \textbf{v}_{i}, \textbf{v}_{r}, \textbf{v}_{j} \right) $ as input and obtain two vectors representing the subject and object, respectively. Function, $g_p$, is used to calculate the relation features vectors of output. It uses an edge vector and two associated object vectors as inputs and outputs a predicted relationship $r$. In the scene graph, given an edge $ \textbf{v}_{r} $, the two associated objects, $ \textbf{v}_{i} $ and $\textbf{v}_{j} $, can be determined. Thus, the output relationship vector $ \textbf{v}_{r}^{\prime}$ can be simply expressed as:
\begin{equation}
    \textbf{v}_{r}^{\prime}=g_p\left(\textbf{v}_{i}, \textbf{v}_{r}, \textbf{v}_{j}\right):\left(o_i, r, o_j\right) \in E
    \label{eq:equ_gp}
\end{equation}

In contrast, the calculation of output object vectors $ \textbf{v}_{i}^{\prime} $, is more complicated. Generally, an object is associated with two or more relations. Therefore, the output vector of an entity $o_i$ is calculated by considering all the vectors directly connected to the object, i.e., $ \textbf{v}_{j} $, and the corresponding relationship vectors, $ \textbf{v}_{r} $. The function $g_s$ in Equation \ref{eq:equ_gs} is used to compute all vectors starting at node $o_i$ and function $g_o$ in Equation \ref{eq:equ_go} is used to compute all vectors ending at node $o_i$. Afterwards, these vectors are collected into $ V_i^{s} $ and $ V_i^{o} $.

\begin{equation}
    V_{i}^{s} = \lbrace g_s \left( \textbf{v}_{i}, \textbf{v}_{r}, \textbf{v}_{j}  \right) : \left( o_i, r, o_j \right) \in E \rbrace
    \label{eq:equ_gs}
\end{equation}

\begin{equation}
    V_{i}^{o} = \lbrace g_o \left( \textbf{v}_{j}, \textbf{v}_{r}, \textbf{v}_{i}  \right) : \left( o_j, r, o_i \right) \in E \rbrace
    \label{eq:equ_go}
\end{equation}

Then the output vector $ \bf{v_i^{'}}$ for the entity $o_i$ is expressed as.

\begin{equation}
    \textbf{v}_{i}^{\prime} = h \left(V_{i}^{s} \cup V_{i}^{o}  \right) \qquad\qquad , 
    \label{eq:equ_h}
\end{equation}
where h denotes a function that pools all vectors in $ \bf{V_i^{s}} $ and $ \bf{V_i^{o}} $ to a single output vector \cite{Johnson2018ImageGF}.

\subsection{Image generator }

\begin{figure}
    \centering
    \includegraphics[width=4in]{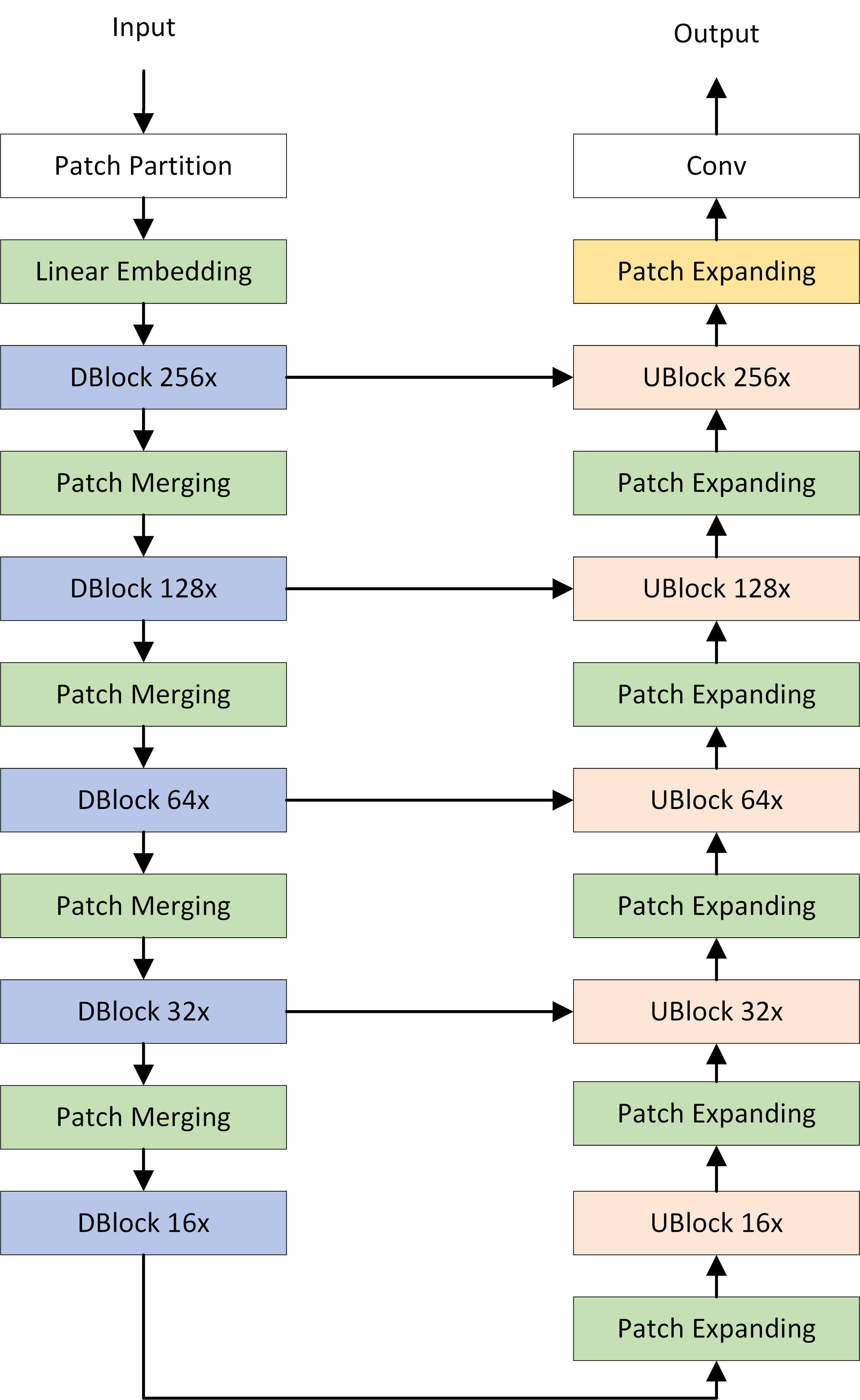}
    \caption{The UNet architecture of the super-resolution submodule.}
    \label{fig:unet}
\end{figure}

The image generator is composed of three diffusion models located downstream. In the diffusion model, a hidden variable $z$ is obtained by adding noise to the image for $T$ times. After forward and backward diffusion, a basic 64 * 64 image can be learned. The basic image is input to the first Swinv2-Unet to generate a 256 * 256 image. Finally, the image goes to the second Swinv2-Unet super-resolution generation, producing a 1024 * 1024 high-definition image. In contrast to Imagen, we focus on improving super-resolution diffusion models. We introduce a new UNet variant to our super-resolution diffusion model, called Swinv2-UNet. The Swin Transformer Block is replaced with the Swin Transformer v2 Block based on the original Swin-Unet \cite{Cao2021SwinUnetUP}, the complete structure of which is shown in Figure \ref{fig:unet}.

The DBlock and UBlock of Swinv2-UNet consist of the Swin Transformer v2 block, which comprises LayerNorm (LN) layers, multi-headed self-attention modules, Residual connections and a 2-layer MLP with GELU non-linearity. The Swin Transformer v2 block could be represented as:

\begin{equation}
    \hat{z} ^ {l+1} = LN(Attn(z^l)) + z^l
    \label{eq:equ_att1}
\end{equation}

\begin{equation}
    z^{l+1} = MLP(LN(\hat{z} ^{l+1}))+ \hat{z} ^{l+1} \quad ,
    \label{eq:equ_att2}
\end{equation}
where $z^l$ and $z^{l+1}$ denote the input and output of the Transformer v2 block, respectively. $\hat{z} ^ {l+1}$ is an intermediate variable. + denotes the residual connection or skip connection. 

The attention of Swin-v2 is expressed as:

\begin{equation}
    Attn(Q, K, V) = SoftMax \left( \frac{Cosine(Q, K)}{\tau} + B\right) \quad ,
    \label{eq:equ_attn_v2}
\end{equation}
where $Q$, $K$, and $V$ denote the matrix of query, key and value, respectively. $Cosine()$ refers to a function that calculates the scaled cosine similarity of $Q$ and $K$. $\tau$ denotes a learnable scalar, usually greater than 0.01. $B$ is a matrix of relative position bias.

\begin{figure}[!h]
    \centering
    \includegraphics[width=\textwidth]{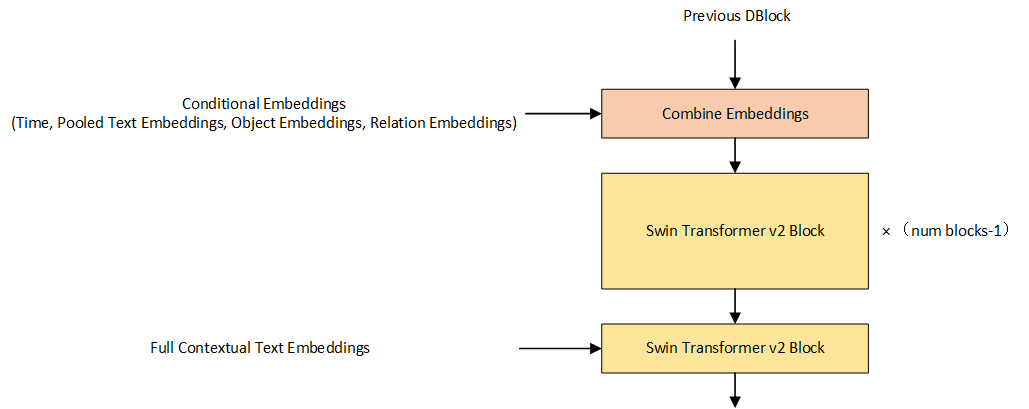}
    \caption{Swinv2-Unet DBlock}
    \label{fig:dbloc}
\end{figure}

Figure \ref{fig:dbloc} illustrates the network structure of the Swinv2-Unet DBlock, which is the basic component of the downsampling path under the encoding-decoding structure of UNet. Firstly, the DBlock combines the pooled text embeddings, object embeddings and relation embeddings into a conditional embedding input to the cross-attention layer. Next, it is followed by the Swinv2-Transformer v2 blocks for (num\_block-1) times feature extraction.

\begin{figure}[!h]
    \centering
    \includegraphics[width=0.8 \textwidth]{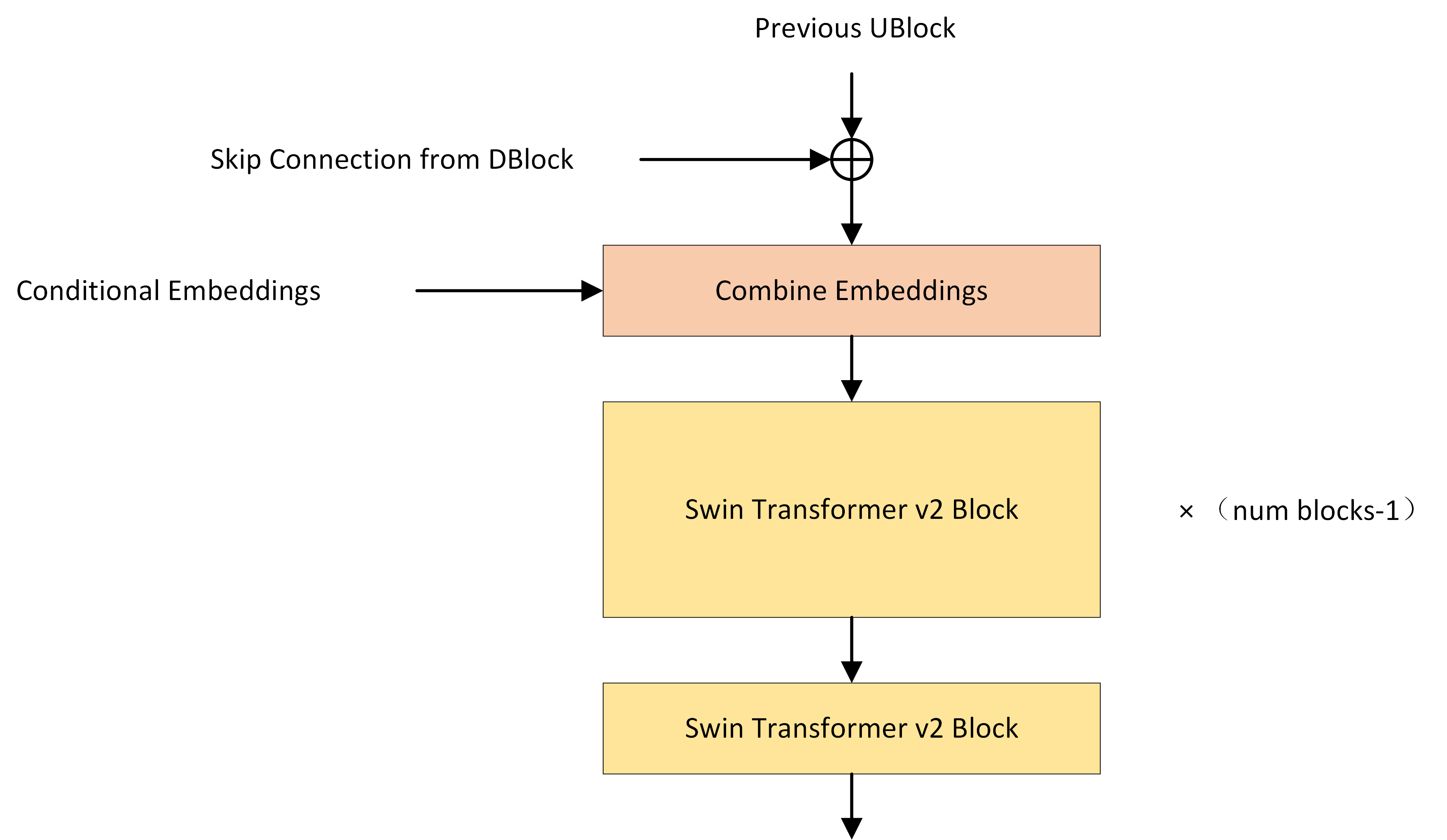}
    \caption{Swinv2-Unet UBlcok}
    \label{fig:ublock}
\end{figure}

Figure \ref{fig:ublock} shows the network structure of the Swinv2-Unet UBlock, which is the basic component of the upsampling path on the UNet encoder-decoder. The inputs to the UBlock include the output of the previous UBlock layer and the corresponding DBlock. The DBlock and UBlock are connected using skip connections \cite{Zhou2020UNetRS}. Subsequently, the conditional embedding inputs are also introduced to the cross-attention layer. Similar to the DBlock, this layer is followed by the Swinv2-Transformer v2 blocks for (num\_block-1) times feature extraction.

The encoder is presented as a stacking of DBlocks and Patch Merging. In the encoder, images are fed into five consecutive DBlocks for learning, where the feature dimension and resolution are maintained. Meanwhile, Patch Merging performs Token Merging and increases the feature dimension to four times the original dimension. Next, we apply a linear layer to standardise the feature dimension to twice the original dimension. The process is repeated four times in the encoder.

Similar to UNet, skip connections are used to integrate the multi-scale features of the encoder with the upsampled features. We connect shallow and deep features to minimise the loss of spatial information due to downsampling. The next layer is a linear layer where the dimensionality of the connected features is kept the same dimensionality as that of the upsampled features.

The decoder is a symmetric decoder corresponding to the encoder. For this reason, unlike the Patch Merging used in the encoder, we use Patch Expanding in the decoder to upsample the extracted features. The Patch Expanding reshapes the feature maps of adjacent dimensions into a higher resolution feature map (2 × upsampling) and accordingly reduces the number of feature dimensions to half the original dimensionality.

\section{Experiments}
In this section, we perform extensive experiments to evaluate the proposed Swinv2-Imagen model by using the MSCOCO, CUB and Multi-modalCelebA-HQ (MM CelebA-HQ) datasets. Firstly, a brief description of the datasets is given. Secondly, we compare the performance of the Swinv2-Imagen model with state-of-the-art generative models. Finally, we conduct ablation experiments to compare the contributions of each module.

\subsection{Setup}
\subsubsection{Datasets}
The Microsoft Common Objects in Context 2014 (MS COCO-2014) \cite{Cho2014LearningPR}, the Caltech-UCSD Birds-200-2011 (CUB-200-2011) \cite{Ho2022ClassifierFreeDG}  and MM CelebA-HQ \cite{Xia2021TediGANTD} datasets are utilised in this research. Three datasets cover both simple (CUB) and complex (MSCOCO) datasets. The use of the MM CelebA-HQ dataset is mainly because most generative models such as Cogview and Craiyon, produce distorted and less realistic faces. 

\begin{itemize}
    \item MSCOCO \footnote{https://paperswithcode.com/dataset/coco} was released in 2014. It is a collection of 164K images, which have been partitioned into the training set (82K), validation set (41K) and testing set (41K). The dataset is complex because most of the images possess at least two objects.
    \item CUB \footnote{https://paperswithcode.com/dataset/cub-200-2011} contains 12K bird images of 200 subcategories, 6K for training and 6K for testing. It is a simple dataset, having only one object per image.
    \item MM CelebA-HQ \footnote{https://paperswithcode.com/dataset/multi-modal-celeba-hq-1} is a large-scale face image dataset. It is a collection of 30K high-resolution face images. The dataset is used widely to train and evaluate algorithms for text-image generation and text-guided image manipulation. 
\end{itemize}

\subsubsection{Evaluation metrics}
We adopt Fréchet Inception Distance (FID) \cite{Heusel2017GANsTB} and Inception Score (IS) \cite{Li2019ObjectDrivenTS} as evaluation metrics. Both are acknowledged as standard metrics for evaluating the image generation model. Specifically, IS examines both the clarity and diversity of the resulting images. The higher the IS, the better the quality of the generated images. FID calculates the difference between the generated image and the original image. The smaller the difference, the better the generated image is.

\subsubsection{Baselines}
\begin{itemize}
    \item PCCM-GAN \cite{Qi2021PCCMGANPT} (Photographic Text-to-Image Generation with Pyramid Contrastive Consistency Model) is a typical multi-stage generative model. Its main innovations include the introduction of stack attention and the lateral connection of the PCCM. The two modules enhance the generative model to simultaneously extract semantic information from both global and local aspects, ensuring that the generated images are semantically consistent.
    \item DM-GAN \cite{Zhu2019DMGANDM} (Dynamic Memory Generative Adversarial Networks for Text-to-Image Synthesis) is also a multi-stage generative model. It uses a memory module and a gate mechanism in the image refinement process. The aim is to re-extract important information from the image as an aid when the generated image is not as good as expected.

    \item SDGAN \cite{Zhang2021CrossModalCL} (Semantics Disentangling for Text-to-Image Generation) consists of two modules, i.e., Siamese and semantic conditioned batch normalization, to extract high-level and low-level semantic features respectively.

    \item CogView \cite{Ding2021CogViewMT} is based on the Transformer architecture. Its input is a text-image pair. The text and image features are combined and passed to the GPT language model for autoregressive training.

    \item GLIDE \cite{Nichol2022GLIDETP} is a large-scale image generation model based on diffusion models with 3.5 billion model parameters.

    \item DALL-E 2 \cite{Ramesh2022HierarchicalTI} is also based on diffusion models. One of its highlights is the use of a priori model built on the diffusion models. Its inputs are also text and corresponding images. The text is first passed through the priori model and a corresponding image vector is generated. The image is passed through the CLIP module which also generates an image vector to supervise the result of the priori model.

    \item LAFITE \cite{Zhou2022TowardsLT} is a variant of generative adversarial networks. It leverages the CLIP model to extract features from images and text, ensuring text-image consistency.

    \item Imagen \cite{Saharia2022PhotorealisticTD} is a text-to-image synthesising model based on the diffusion model. It passes text through a large pre-trained T5 language model and generates high-fidelity images through cascading diffusion model blocks.
\end{itemize}

\subsubsection{Training parameters }
We apply an Imagen-like training strategy, i.e., training the base model and then the super-resolution model twice. The Adam optimiser is adopted, having a learning rate of 1e-4. We give 10,000 linear warm-up steps with a batch size of 8 and training epochs of 1,000. The loss function is Mean Squared Error (MSE), formulated as follows.

\begin{equation}
    MSE(I, K) = \frac{1}{M \times N} \sum_{i=0}^{M-1} \sum_{j=0}^{N-1} [I(i, j) - K(i, j)]^2 \quad ,
    \label{eq:equ_lossfun}
\end{equation}
where $M$ and $N$ denote the total number of pixels in the real image $I$ and the generated image $K$, respectively. A smaller MSE implies that the generated image is closer to the real image. 

\subsection{Experimental Results}
In this subsection, we evaluate the proposed model by comparing it against a few state-of-the-art generative models.

\subsubsection{Performance evaluation}

\begin{table}[!h]
\centering
\begin{tabular}{|c|c|c|c|c|c|} 
\hline
\multirow{2}{*}{Model} & \multicolumn{2}{c|}{MSCOCO}    & \multicolumn{2}{c|}{CUB}       & MM CelebA-HQ    \\ 
\cline{2-6}
                       & FID \textbf{↓} & IS \textbf{↑} & FID \textbf{↓} & IS \textbf{↑} & FID \textbf{↓}  \\ 
\hline
PCCM-GAN \cite{Qi2021PCCMGANPT}          & 33.59          & 26.52         & 22.15          & 4.65          & -               \\ 
\hline
DM-GAN \cite{Zhu2019DMGANDM}             & 32.64          & 30.49         & 16.09          & 4.75          & 131.05          \\ 
\hline
SDGAN \cite{Zhang2021CrossModalCL}             & 29.35          & 35.69         & -              & 4.64          & -               \\ 
\hline
DALL-E \cite{Zhang2021CrossModalCL}            & 27.5           & 17.9          & 56.1           & -             & 12.54           \\ 
\hline
CogView \cite{Ding2021CogViewMT}           & 13.9           & 18.2          & -              & -             & -               \\ 
\hline
GLIDE \cite{Nichol2022GLIDETP}             & 12.24          & -             & -              & -             & -               \\ 
\hline
DALL-E 2 \cite{Ramesh2022HierarchicalTI}           & 10.39          & -             & -              & -             & -               \\ 
\hline
LAFITE \cite{Zhou2022TowardsLT}            & 8.12           & 32.34         & 10.48          & 5.97          & 12.54           \\ 
\hline
Make-A-Scene \cite{Gafni2022MakeASceneST}      & 7.55           & -             & -              & -             & -               \\ 
\hline
Imagen \cite{Saharia2022PhotorealisticTD}             & 7.27           & -             & -              & -             & -               \\ 
\hline
{\bfseries Swinv2-Imagen}          & {\bfseries 7.21}          & {\bfseries 31.46}         & {\bfseries 9.78}           & {\bfseries 8.44}          & {\bfseries 10.31}           \\
\hline
\end{tabular}
\caption{Experimental results of varied models for Text-To-Image synthesis. Symbols ↑ and ↓ indicate the higher the best and the lower the best, respectively. – means that the indicator is not used in the article.}
\label{tab:evaluation}
\end{table}

Table \ref{tab:evaluation} demonstrates the results of the quantitative comparison. The proposed model is compared against 10 popular generative models, including GAN and diffusion models. It is evident that the proposed Swinv2-Imagen model outperforms the baselines on all three datasets. Particularly, on the MSCOCO dataset, Swinv2-Imagen significantly outperforms the GAN-based generative model and slightly surpasses the Imagen, achieving an FID of 7.21.

\subsubsection{Qualitative analysis}
\begin{figure}
    \centering
    \includegraphics[width=6in]{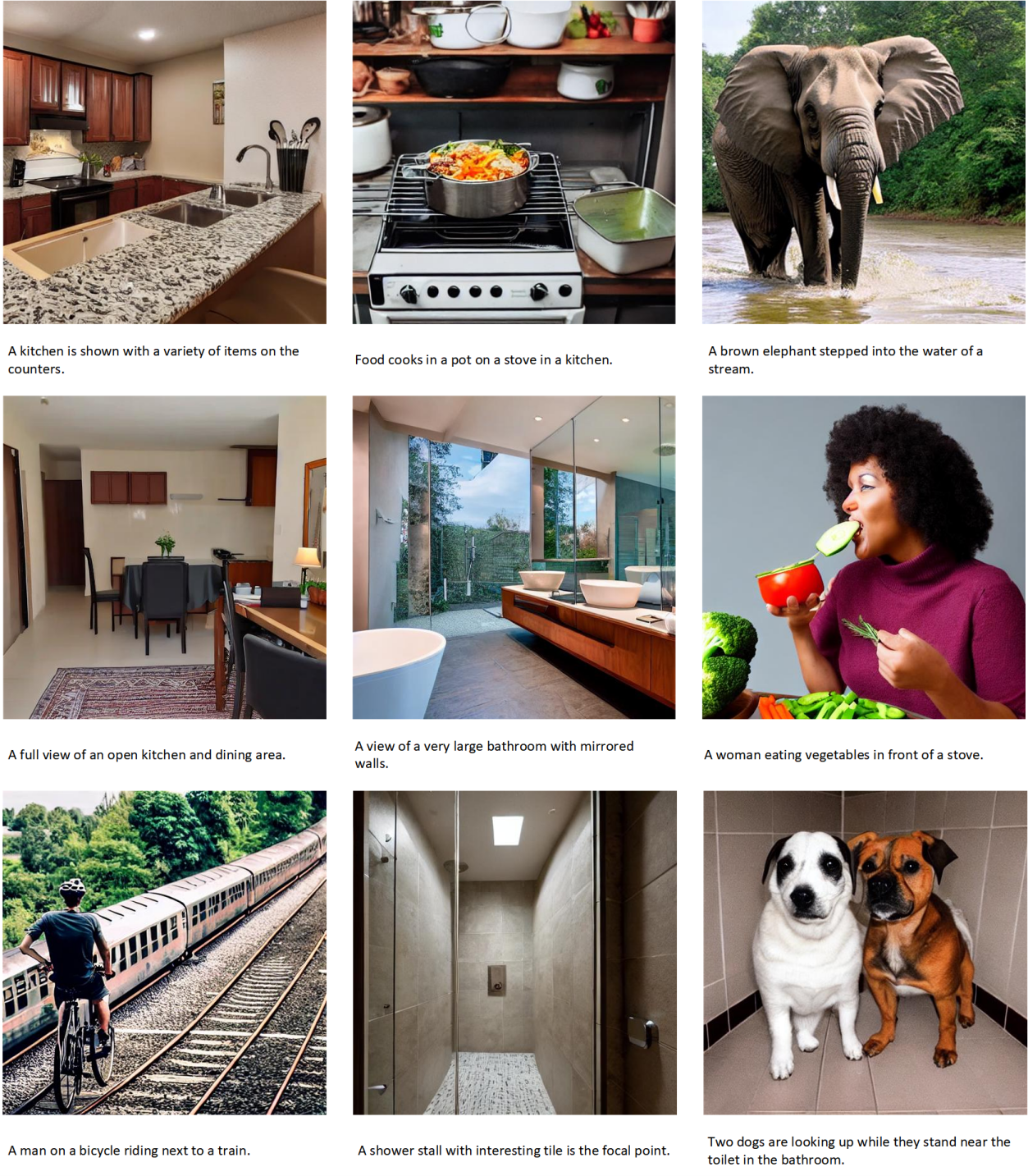}
    \caption{Generated examples on MSCOCO}
    \label{fig:coco}
\end{figure}

\begin{figure}
    \centering
    \includegraphics[width=6in]{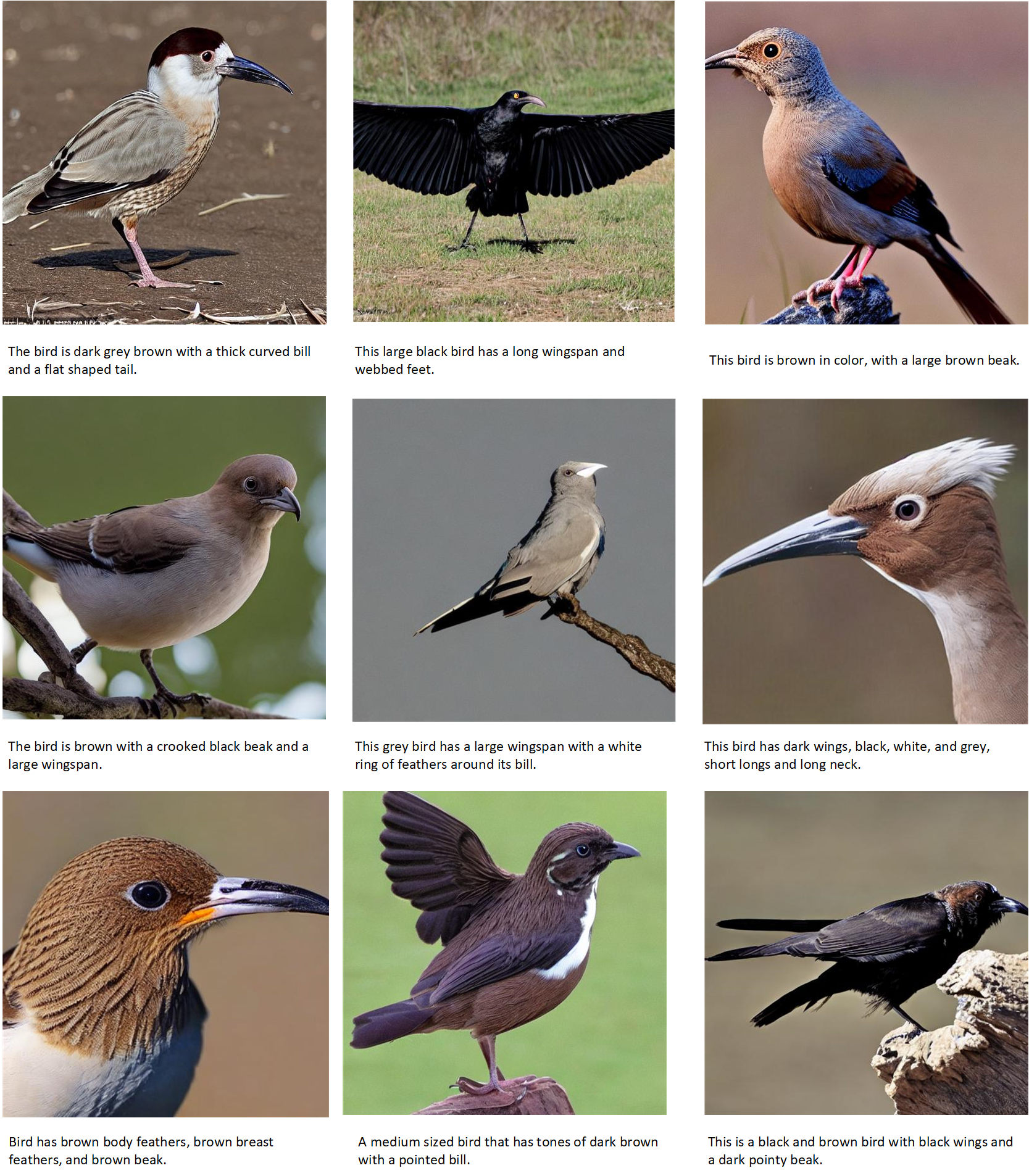}
    \caption{Generated examples on CUB}
    \label{fig:cub}
\end{figure}

\begin{figure}
    \centering
    \includegraphics[width=6in]{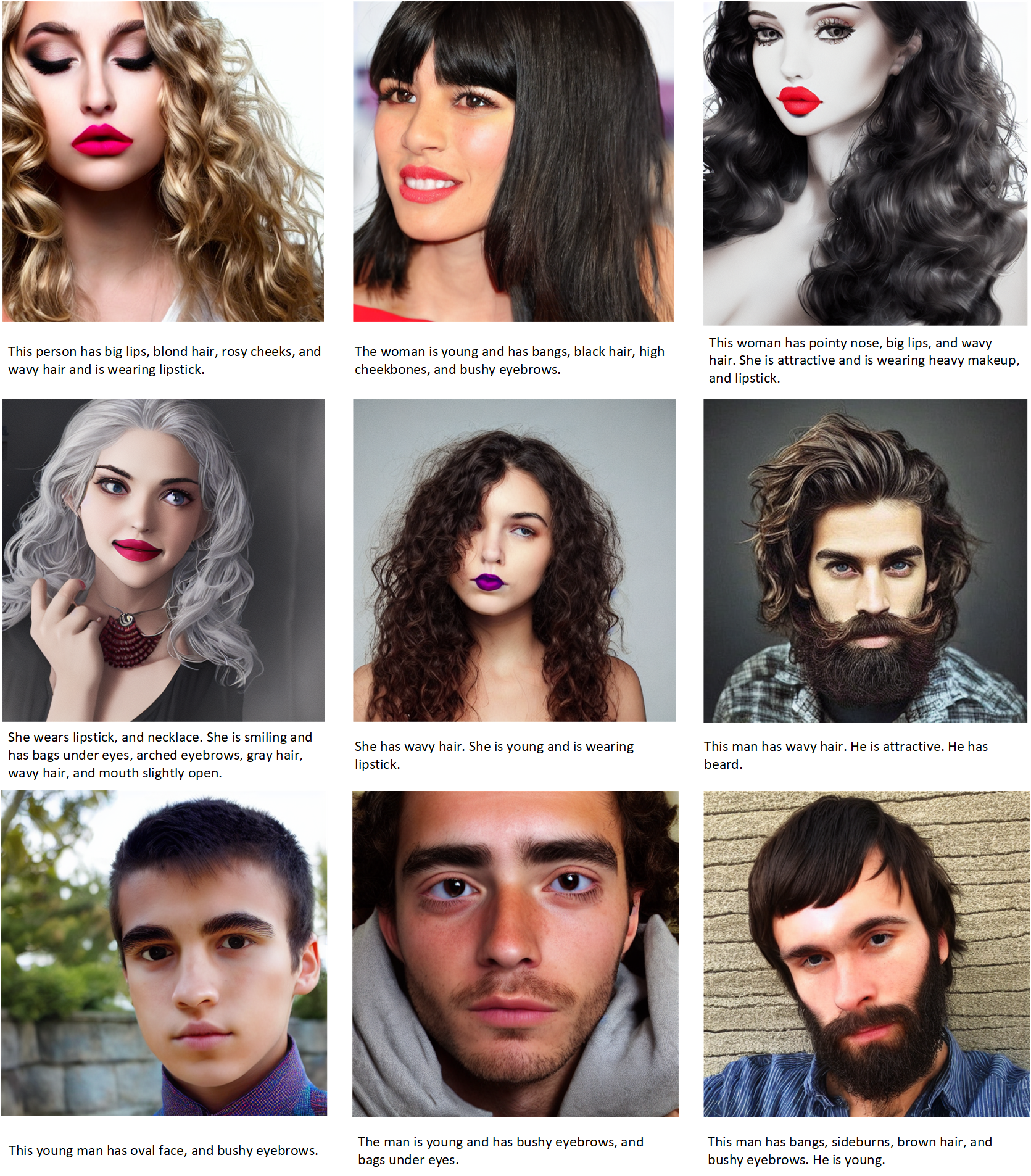}
    \caption{Generated examples on MM CelebA-HQ}
    \label{fig:face}
\end{figure}

Figures \ref{fig:coco}, \ref{fig:cub} and \ref{fig:face} show examples of images generated by our proposed model on MSCOCO, CUB and MM CelebA-HQ, respectively. It can be seen that our model understands the text very well. For example, given the text input, ‘Food cooks in a pot on a stove in a kitchen’, the resulting picture not only contains the food, the stove and the pot, but also places these objects to the exact location. More importantly, based on the word ‘kitchen’, the model also generates other common kitchen objects, such as spoons and storage shelves. This shows that our model understands the text accurately and comprehensively.

\subsection{Ablation study}

In order to improve the performance of the generation models, we introduce two new modules to Imagen, i.e., scene graph and Swinv2-Unet. These are the main innovations of the article. In this subsection, two ablation experiments are conducted on MSCOCO to investigate the contributions of the scene graph module and Swinv2-Unet, respectively. The choice to experiment on MSCOCO is based on two considerations. Firstly, each image in MSCOCO contains multiple objects, which is more complex than CUB dataset. Theoretically, it allows a better evaluation on the effect of each module. Secondly, the main baseline we referenced, Imagen, is only experimented on MSCOCO. The aim of Experiment 1 is to evaluate the contribution of the scene graph module. We add only the scene graph, and the diffusion model is still built using Efficient-Unet, which is called Imagen\_sg. Experiment 2 is designed to evaluate the performance of the Swinv2-Unet. We constructed a new diffusion model using our improved Swinv2-Unet and replace Imagen's super-resolution diffusion models with it, which is called Swinv2-Imagen\_su. The result of Experiment 1 supports our conjecture that merely using a T5 encoder does not sufficiently learn the semantic information of the text, as mentioned in the introduction. Experiment 2 shows that the diffusion model constructed with the Transformer outperforms the CNN-constructed diffusion model in the image generation task. It also can be seen from Table \ref{tab:ablationstudy} that the FIDs of the Imagen\_sg and Swinv2-Image\_su are very close. This intuitively reveals that the two submodules almost contribute equally to the FID.

\begin{table}[!h]
    \centering
    \begin{tabular}{|c|c|c|c|}
    \hline
    \textbf{Model} & \textbf{Scene Graph} & \textbf{Swinv2-UNet} & \textbf{FID}\\
    \hline
     Imagen & ~ & ~ & 7.27 \\
     Imagen\_sg & YES & ~ & 7.24 \\
     Swinv2-Imagen\_su & ~ & YES & 7.23 \\
     Swinv2-Imagen & YES & YES & 7.21 \\
     \hline
    \end{tabular}
    \caption{Ablation study of Swinv2-Imagen model}
    \label{tab:ablationstudy}
\end{table}

\section{Conclusion}
In this paper, we propose a novel text-to-image synthesis model based on Imagen, called the Swinv2-Imagen, which integrates the Transformer and Scene Graph. The improved sliding window-based hierarchical visual Transformer (Swin Transformer v2) avoids the local view of CNN convolution operations. It improves the efficiency and effectiveness of the Transformer applied to image generation. In addition, we introduce a Scene Graph in the text processing stage. Feature vectors of entities and relationships are extracted from the Scene Graph and incorporated into the diffusion model. These additional feature vectors improve the quality of generated images. Swinv2-Imagen produces 1024 × 1024 samples with unprecedented fidelity with these novel components.

Furthermore, it has also recently been noted that autoregressive models can produce diverse and high-quality images from text. Thus, we plan to consider combining autoregressive and diffusion models for image generation and determine the best opportunities to combine their strengths.

\bibliographystyle{unsrt}  
\bibliography{main}

\end{document}